\documentclass[letterpaper, 10 pt, conference]{ieeeconf}  
\IEEEoverridecommandlockouts                              
\usepackage{cite}
\usepackage{amsmath,amssymb,amsfonts}
\usepackage{graphicx}
\usepackage{textcomp}
\usepackage{xcolor}

\usepackage[ruled,lined]{algorithm2e}
\usepackage{algpseudocode}

\usepackage{tabu}                     
\usepackage{multirow}                 
\usepackage{multicol}                 
\usepackage{float}                    
\usepackage{makecell}                 
\usepackage{booktabs}                 
\def\BibTeX{{\rm B\kern-.05em{\sc i\kern-.025em b}\kern-.08em
    T\kern-.1667em\lower.7ex\hbox{E}\kern-.125emX}}

\title{\LARGE \bf
Multi-directional Safe Rectangle Corridor-Based MPC for Nonholonomic Robots Navigation in Cluttered Environments*
}

\author{Yinsong Qu, Yunxiang Li and Shanlin Zhong 
\thanks{*This work is supported by the STI 2030—Major Projects 2021ZD0200400, National Natural Science Foundation of China (Grant No.62203443), Major program of the National Natural Science Foundation of China (Grant No. T2293720/3724/3725), the Excellent Youth Program of State Key Laboratory of Multimodal Artificial Intelligence Systems (E4SP100101). (Corresponding author: Shanlin Zhong) }
\thanks{ Shanlin Zhong is with The State Key Laboratory of Multimodal Artificial Intelligence Systems, Institute of Automation, Chinese Academy of
Sciences, Beijing, 100190, China.
        (phone:+8601082544702; email: shanlin.zhong@ia.ac.cn)}
\thanks{Yinsong Qu and Yunxiang Li are with School of Automation, Beijing Institute of Technology, Beijing 100081, China
        (email: 3120246387@bit.edu.cn; 3120235332@bit.edu.cn)}
}

\begin{document}

\maketitle
\thispagestyle{empty}
\pagestyle{empty}

\begin{abstract}
Autonomous Mobile Robots (AMRs) have become indispensable in industrial applications due to their operational flexibility and efficiency. Navigation serves as a crucial technical foundation for accomplishing complex tasks. However, navigating AMRs in dense, cluttered, and semi-structured environments remains challenging, primarily due to nonholonomic vehicle dynamics, interactions with mixed static/dynamic obstacles, and the non-convex constrained nature of such operational spaces. To solve these problems, this paper proposes an Improved Sequential Model Predictive Control (ISMPC) navigation framework that systematically reformulates navigation tasks as sequential switched optimal control problems. The framework addresses the aforementioned challenges through two key innovations: 1) Implementation of a Multi-Directional Safety Rectangular Corridor (MDSRC) algorithm, which encodes the free space through rectangular convex regions to avoid collision with static obstacles, eliminating redundant computational burdens and accelerating solver convergence; 2) A sequential MPC navigation framework that integrates corridor constraints with barrier function constraints is proposed to achieve static and dynamic obstacle avoidance. The ISMPC navigation framework enables direct velocity generation for AMRs, simplifying traditional navigation algorithm architectures. Comparative experiments demonstrate the framework's superiority in free-space utilization ( an increase of 41.05$\%$ in the average corridor area) while maintaining real-time computational performance (average corridors generation latency of 3 ms). 

\end{abstract}

\section{Introduction}

\begin{figure}[htbp]
\centering
\includegraphics[width=0.9\linewidth]{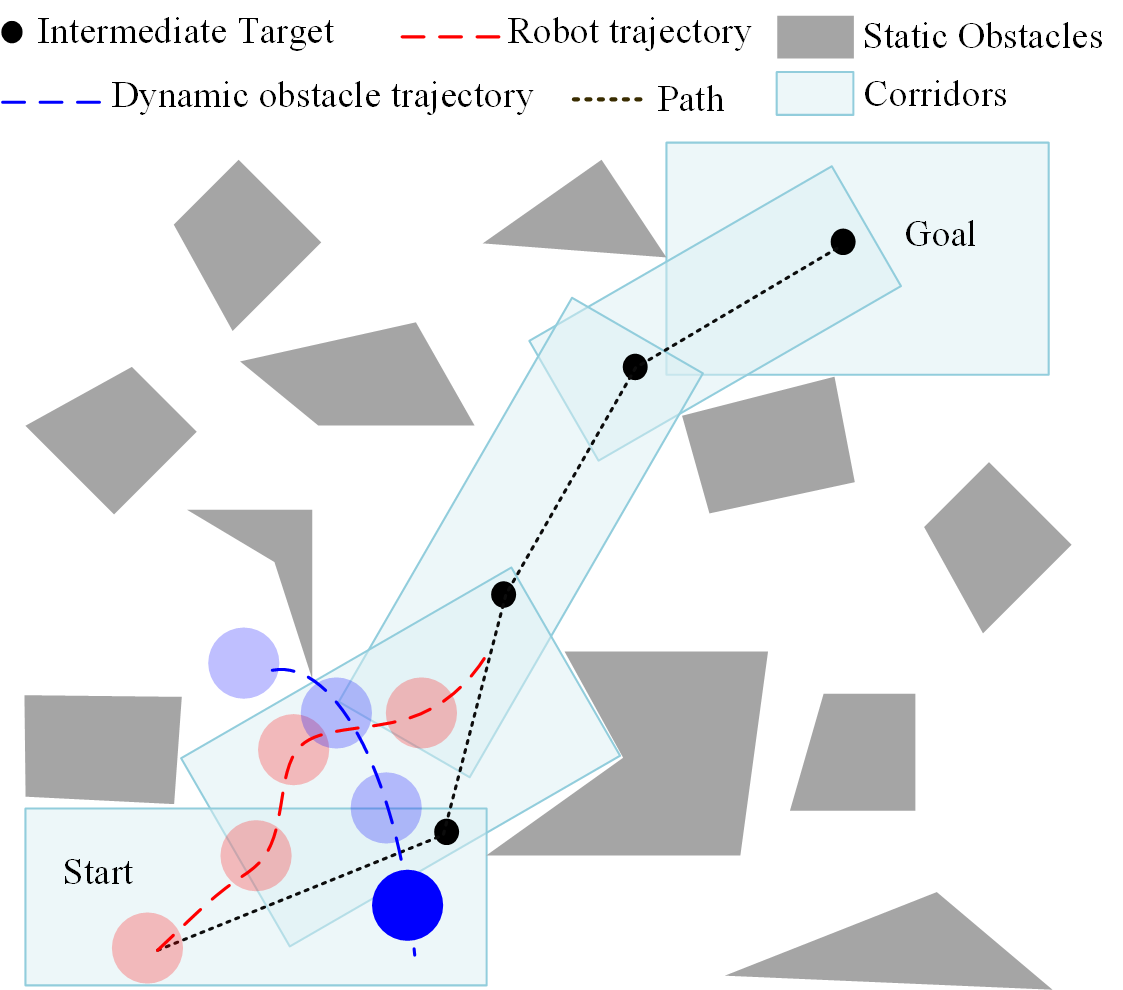}
\caption{Schematic diagram of ISMPC based on MDSRC constraints. The robot navigation problem is reformulated as a switching optimal control problem. By continuously switching corridor constraints and intermediate targets, the robot reaches the navigation goal. In each corridor, it is only constrained by the corridor boundaries and dynamic obstacles, reducing constraint complexity and easing optimal control problem solving.}
\label{fig1}
\end{figure}

Automated Mobile Robots (AMRs) are widely used in industrial and household service scenarios, where work environments are complex and diverse, with various constraints from static and dynamic obstacles [1]. Therefore, safe navigation to the target point is the basis for completing different tasks [2]. Currently, there is a significant amount of research on safe navigation of mobile robots. Most research generally separates trajectory planning from trajectory tracking, assuming that the trajectory tracking controller can perfectly follow the planned trajectory [3]. However, this assumption is not satisfied all the time due to the existence of disturbances. Trajectory planning can be further divided into two parts: discrete path search and continuous trajectory optimization. To reduce the constraint equations in the optimization part, a concept called the flight corridor has been proposed in [4][5]. This method represents the feasible region for robots in unstructured environments with a series of convex regions that do not contain obstacles, thereby constructing the constraint conditions for the optimization problem [6]. The advantage of this method is that no matter how many obstacles there are, they can always be separated by a small number of convex regions, which can greatly reduce the constraint conditions. The current methods face the following issues: (1) The time consumption of front-end path search and corridor construction leads to the difficulty to avoid dynamic obstacles in real time. (2) It cannot ensure a high utilization rate of the free space. (3) The separation of trajectory planning and trajectory tracking can easily lead to tracking issues and result in collision problems.

To solve the above problems, the sequential MPC scheme is used to generate the desired velocity of the AMR, integrating both trajectory planning and trajectory tracking. Drawing on the concept of the safe flight corridor, the Multi-Directional Safety Rectangular
Corridor (MDSRC) algorithm is proposed to construct constraints of MPC. Meanwhile, the dynamic control barrier functions (D-CBFs) are used to avoid dynamic obstacles in real-time.

\subsection{Related Work}
\textit{Motion Planning}: Robot motion planning methodologies broadly comprise graph search [7][8], probabilistic sampling [9][10], potential/vector fields [11][12], optimal control (OCP) [13]--[17], and learning-based approaches [18][19]. While graph search and probabilistic sampling yield sparse paths requiring subsequent trajectory optimization, learning-based methods suffer from interpretability limitations and data-intensive training demands [15]. Potential/vector field methods synthesize obstacle repulsion and target attraction fields but risk local minima and control oscillations. OCP-based methods uniquely reconcile kinematic constraints, obstacle avoidance, and trajectory smoothness.

Recent OCP implementations leverage safe traffic corridors (STCs): [13][14] optimize time-minimal trajectories within corridors, while [15] exploits differential flatness to bypass motion model discretization issues. These approaches assume perfect trajectory tracking – an idealized condition complicating practical deployment. Model Predictive Control (MPC) addresses this limitation by unifying planning and control [16][17]. Building on sequential MPC frameworks [17], our navigation scheme innovates through dual intermediate target integration within the MPC formulation, reducing computational latency while maintaining STC-based obstacle constraints.

\textit{Corridor Construction}: The Semidefinite Programming-based Iterative Regional Inflation (IRIS) method [20] partitions free space into convex regions by iteratively expanding obstacle-free seed points, though its reliance on mixed-integer programming induces high computational complexity, limiting real-time applicability. To address this, octree-based tetrahedral decomposition [21] and path-guided 
tetrahedral inflation [22] accelerate convex decomposition but underutilize free space. Polyhedral corridor inflation [22] prioritizes computational efficiency (2--3 ms generation) over optimality, producing narrower corridors than IRIS. Drawing inspiration from the aforementioned 3D corridor constraint methods for UAVs, planar rectangular inflation method for AMRs is proposed in [13] and [14] to simplify corridor construction process and reduce computation time. However, this approach exhibits poor free-space approximation in unstructured environments. Building upon this foundation, our proposed multi-directional rectangular inflation method significantly improves free-space utilization by expanding directional degrees of freedom, effectively resolving previous limitations in geometric adaptability.

\subsection{Motivations, Contributions, and Organization}
This paper aims to solve the navigation problem in the cluttered environment with mixed static/dynamic obstacles. An sequential model predictive control framework based on MDSRC and D-CBF is proposed to tackle static and dynamic obstacles separately (Figure 1), reducing constraints count of MPC while ensuring collision avoidance. The contributions of this paper are as follows.

(1) A multi-directional dynamically expandable rectangular corridor construction method is proposed to optimize static obstacle constraint representation through adaptive spatial partitioning, simultaneously enlarge individual corridor areas, and reduce the total constraint count. 

(2) A sequential MPC navigation framework that integrates rectangular corridor constraints with barrier function constraints is proposed to achieve static and dynamic obstacle avoidance. Meanwhile, a dual - target cost function is designed to reduce trajectory length.

(3) Simulation experiments conducted in Gazebo validate the effectiveness of the proposed MDSRC-based ISMPC navigation scheme.

The paper is structured as follows. Section II formulates the nonholonomic MPC optimization problem, Section III details the ISMPC-MDSRC co-design methodology, Section IV presents comparative experimental validation, and Section V concludes this paper.

\section{Problem Formulation}
Assuming the mobile robot can be controlled by velocity, only the kinematic model of the robot needs to be considered, and the kinematic model without side slip motion is given as follows.
\begin{equation*}
\boldsymbol{\dot{\eta}}= \begin{bmatrix}\cos\psi& 0\\ \sin\psi& 0\\ 0& 1\end{bmatrix}\boldsymbol{\nu}.  \tag{1}
\end{equation*}
where $\boldsymbol{\eta}=[\boldsymbol{\bar\eta}^T,\psi]^T$ represents the position and orientation vector of the robot in the world coordinate system, $\boldsymbol{\bar\eta}=[x,y]^T$, $\boldsymbol{\nu}=[u,v,r]^T$ denotes the velocity vector of the robot. Equation (1) can be simply stated as follows:
\begin{equation*}
\boldsymbol{\dot{\eta}}= \boldsymbol{f}(\boldsymbol{\eta},\boldsymbol{\nu})  \tag{2}
\end{equation*}


The whole workspace of a mobile robot is defined as a two-dimensional Euclidean space, $\mathcal{C}= \mathcal{C}_o \cup \mathcal{C}_{free} \in \mathbb{R}^{2}$, where $\mathcal{C}_o$ is the space occupied by obstacles and $\mathcal{C}_{free}$ is the free space without obstacles. The objective of robot navigation is to move a robot from the current position to a desired predefined position while avoiding collisions with static and moving obstacles in the environment. Combined with the MPC scheme, the navigation problem can be solved by a sequential optimization process until the robot reaches the goal. At current time t, the optimization problem is formulated as [17]:
\begin{align}
\min\limits_{\boldsymbol{\nu}_{k:k+N-1}} & \quad J(t)=\boldsymbol{J}_N(\boldsymbol{\eta}_{k+N\vert k})+ \sum\limits_{k=0}^{N-1}\boldsymbol{J}_k(\boldsymbol{\eta}_{k}, \boldsymbol{\nu}_{k})\tag{3a} \\
\quad\quad\mathrm{s}.\mathrm{t}. 
&\quad\boldsymbol{\eta}_{k+1}=\boldsymbol{g}(\boldsymbol{\eta}_{k},\boldsymbol{\nu}_{k}), k=0, \ldots, N-1\tag{3b}\\ 
&\quad\boldsymbol{\nu}_{k}\in \mathcal{U}, k=0, \ldots, N-1 \tag{3c}\\
&\quad\boldsymbol{\bar\eta}_{k}\in \mathcal{C}_{free}, k=0, \ldots, N\tag{3d}\\
&\quad\boldsymbol{\eta}_{0}= \boldsymbol{\eta}_{t},\tag{3e}
\end{align}
where $\mathbf{g}(\boldsymbol{\eta}_{k},\boldsymbol{\nu}_{k}) = \Delta T f(\boldsymbol{\eta}_{k}, \boldsymbol{\nu}_{k})+\boldsymbol{\eta}_{k}$, $\boldsymbol{\bar\eta}_{k}=[x_k,y_k]^T$, $\Delta T$ is the sample time of MPC, $\mathcal{U}$ is the velocity constraints set, $\boldsymbol{\eta}_t$ is the current states of mobile robot, $\boldsymbol{J}_N(\boldsymbol{\eta}_{k+N\vert k})$ and $\boldsymbol{J}_k(\boldsymbol{\eta}_{k}, \boldsymbol{\nu}_{k})$ are the terminal cost function and process cost function expressed by the 
Euclidean distance between the robot and the target point.  The optimization problem (3a)$\sim$(3e) is solved real time, and the first optimized variable in  $\boldsymbol{\nu}_{k:k+N-1}$  is chosen as the control input to force the mobile robot to move to the navigation goal. 


To formulate the model predictive control (MPC) problem, real-time acquisition of system states and constraint parameters is essential. The robotic system's current states $\boldsymbol{\eta}_t$ can be directly measured through odometry, while the input constraint set $\mathcal{U}$ is obtainable via systematic identification procedures. The primary challenge involves constructing an effective collision-free configuration space $\mathcal{C}_{free}$. The subsequent section details the methodology for establishing collision-avoidance constraints, which are systematically formulated as two complementary components: static corridor constraints and dynamic control barrier function constraints. These respective constraint formulations will be comprehensively discussed in Sections III.A (Static Corridor Constraints) and III.B (Dynamic Control Barrier Function Constraints).

\section{Sequential MPC Based Navigation}
The navigation environment is modeled as a 2D occupancy grid map, where static obstacles (e.g., walls, furniture, pillars) prevail over sparse dynamic obstacles (primarily pedestrians) in typical indoor/industrial settings.To achieve computationally efficient constraint handling, our framework implements an environmental decomposition strategy. Static constraints are generated through safety corridor-based convex regions, eliminating the need for per-obstacle collision checks, while dynamic constraints employ D-CBFs to confine control inputs within rigorously certified safe domains, ensuring real-time collision avoidance. The dual-constraint architecture ensures both computational tractability and strict safety guarantees across heterogeneous obstacle types.

\subsection{Static Corridors Constraints}\label{AA}
To construct the convex regions of free space, conventional approaches employ parallel-aligned rectangular regions for incremental spatial partitioning. However, as demonstrated in Figure 2, such axis-parallel constructs may yield insufficiently small regions to enable feasible path navigation in geometrically constrained environments. Notably, oblique-axis construction strategies (e.g., 45-degree orientation in Figure 2) generate significantly larger admissible regions compared to traditional axis-aligned methods. To optimize region coverage, we propose an adaptive multi-directional construction paradigm: (1) Generate candidate regions across $N_c$ distinct angular configurations at each way point $P_i$; (2) Select the maximum-volume region as the safety corridor.

\begin{figure}[htbp]
\centering
\includegraphics[width=0.6\linewidth]{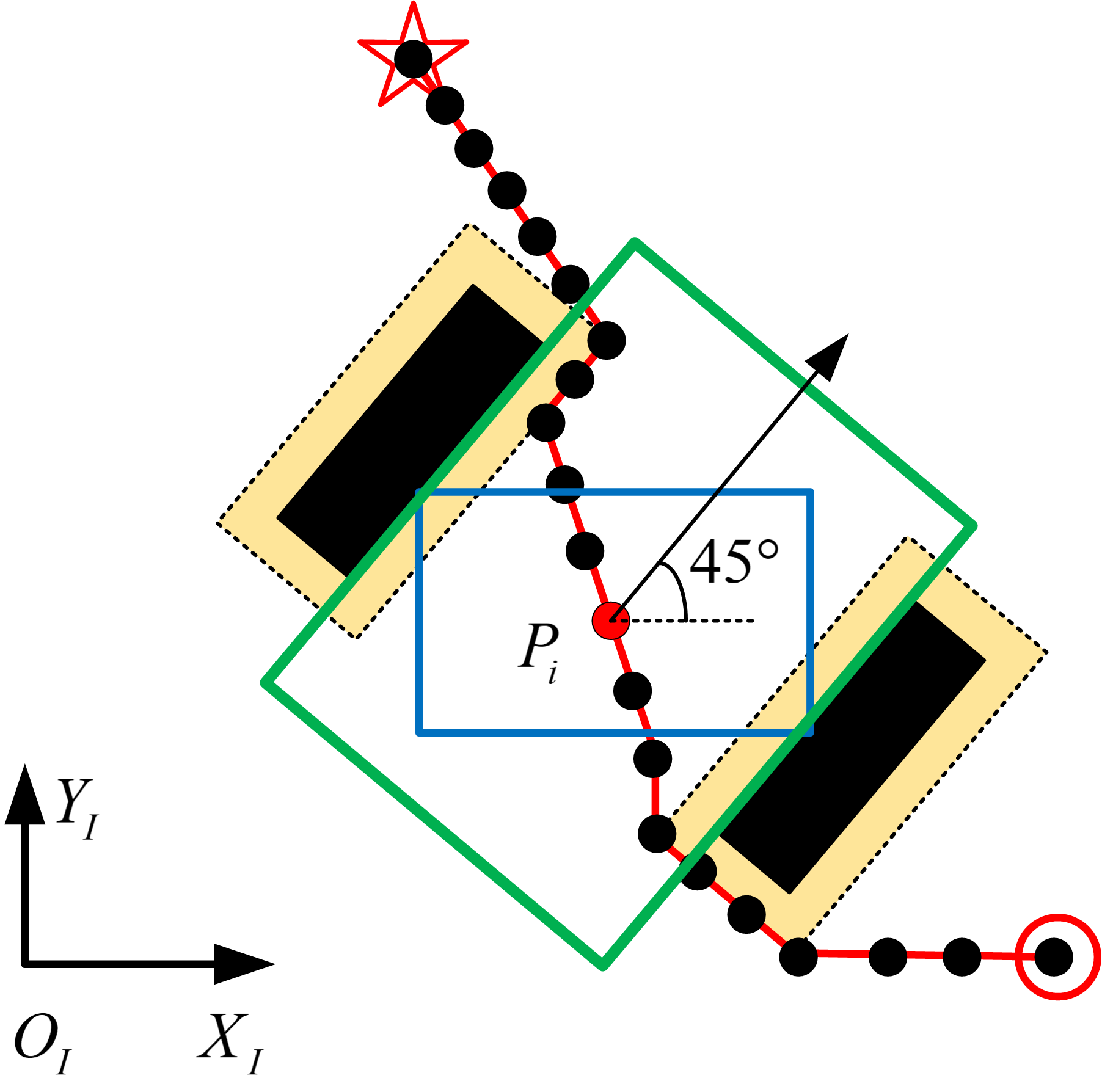}
\caption{Discrete Path Points and Corridor in point $P_i$.}
\label{fig1}
\end{figure}

\begin{figure}[htbp]
\centering
\includegraphics[width=0.6\linewidth]{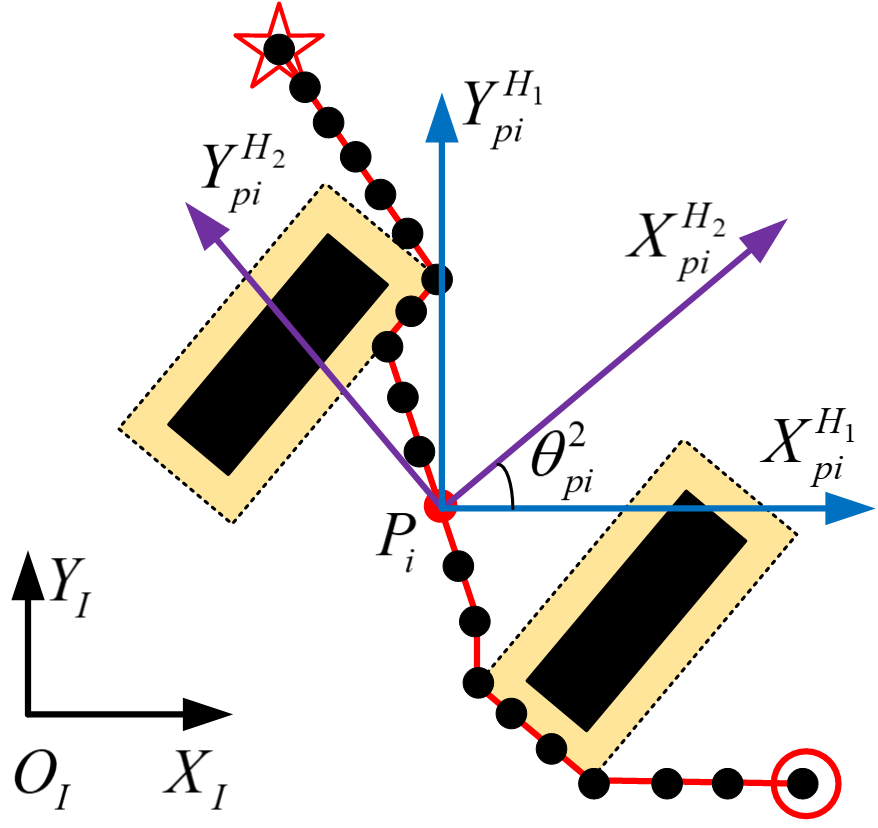}
\caption{The local coordinate systems $H_k$ in $P_i$.}
\label{fig2}
\end{figure}

\begin{figure}[htbp]
\centering
\includegraphics[width=0.6\linewidth]{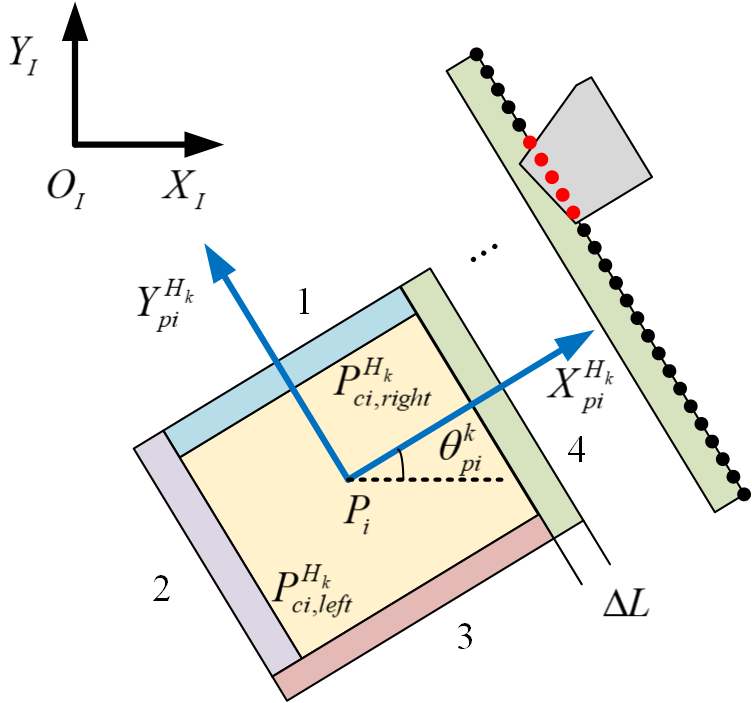}
\caption{Single direction expansion in local frame $H_k$.}
\label{fig3}
\end{figure}

To formalize this angularly-adaptive framework, we define a set of local coordinate systems $H_k$ ($k = 1, ..., N_c$) at each $P_i$, which is shown in Figure 3, where the orientation angle $\theta_{pi}^k$ is calculated as follows.
\begin{align}
\theta_{pi}^{k}&= 90^o*(k-1)/N_c, k=1,...,N_c  \tag{4}
\end{align}
where the primary system ($k = 1$) aligns with the global inertial frame, while subsequent systems ($k \ge 2$) progressively rotate about the z-axis. This parametric formulation enables systematic exploration of orientation-dependent region geometries while maintaining mathematical tractability.

To precisely delineate the rectangle construction process, point 
$P_i$ is designated as the seed point for rectangle generation. The methodology for generating the maximum area rectangle within the local coordinate system is adapted from [13]. The construction process of the rectangular corridor for point $P_i$ is illustrated in Figure 4, with a detailed procedure outlined in Algorithm 1. Departing from [13], this study generates rectangular corridors in multiple directions (Figure 4) for the current path point, selecting the largest rectangular area as the corridor for that path point. Consequently, the $CorridorInfo$ in Algorithm 1 encompasses not only the path point $P_i$ and the coordinates of two vertices but also the rotation angle $\theta_{pi}^{k}$ of the local coordinate system $H_k$ for corridor expansion. The algorithm proceeds as follows: (1) Select $N_c$ local coordinate systems for the corridor based on the current path point, adhering to the selection rules illustrated in Figure 3 and Equation 4. (2) In each coordinate system, expand the rectangular area in the four directions depicted in Figure 4 until the expansion length in the current direction exceeds the predefined maximum value $L_{max}$ or becomes invalid (due to collision with obstacles or exceeding map boundaries), resulting in $N_c$ corridor areas. (3) Compute the area of each corridor region, select the largest area as the corridor region for the current path point, and output the $CorridorInfo$.

Algorithm 1 incorporates two critical computational modules for geometric validation:

(1) Vertex Order Standardization:
The function $updateVertex(P_{temp})$ returns an array $V^{H_k}_{ccs}\in \mathbb{R}^{2\times5}$,  which includes the four vertices of the rectangle in reverse order (with the first vertex included).

(2) Directional Validity Verification:
The function $flag_{valid}=IsValid(P_i,V^{H_k}_{Css}[j],V^{H_k}_{Css}[j+1],\theta_{pi}^{k}, \mathcal{M})$ performs a rigorous spatial validation through:

(i) Coordinate Transformation:
For each vertex pair $(V^{H_k}[j],V^{H_k}[j+1])$ in local frame $\{H_k\}$, apply:
\begin{equation}
V^{I}[j] = P_i + \mathbf{R}(\theta_{pi}^{k})V^{H_k}[j] \tag{5}
\end{equation}
where $\mathbf{R}(\theta_k) \in SO(2)$ denotes the rotation matrix from $\{H_k\}$ to global inertial frame $I$, $\theta_{pi}^{k}$ is the rotation angle of the $k$th local frame $\{H_k\}$, $P_i$ is the $i$th path point's global coordinates, $V_I[j]$ is the $j$th vertex coordinate of the $k$th rectangle corridor in frame $\{I\}$.

(ii) Dual Boundary-Obstacle check:
Validate the transformed segment $[V^{I}[j],V^{I}[j+1]]$ via:
\begin{align}
&V^{I}[j] \in \mathcal{M}, V^{I}[j+1] \in \mathcal{M},\forall j\in\{1,2,3,4\} \tag{6} \\
&P_{j,s} = V^{I}[j] + (s-1)\frac{\mathbf{n}_jL_j}{N_l} \in \mathcal{C}_{\text{free}} , \forall s \in \{1,...,N_l\}\tag{7}
\end{align}
where (6) ensures map boundary compliance, and (7) certifies obstacle-free status through $N_l$ sampled points along the segment normal direction $\mathbf{n}_j$ with step size $L_j/N_l$.

Algorithm 1 outlines the process for generating a corridor from a single path point, while the construction of a corridor sequence containing $N$ path points is detailed in Algorithm 2. This algorithm is straightforward and involves a loop that calls Algorithm 1 for each point on the path to generate its corresponding corridor. To reduce the number of corridors, the algorithm skips a path point and does not generate a corridor for it if the next path point lies within the previous corridor. Additionally, to ensure that adjacent corridors share intersecting point sets, the path point immediately preceding the one not located in the previous corridor is selected as the seed point for the new corridor, as indicated in line 6 of Algorithm 2: $i = i - 1$.

\begin{algorithm}
    \caption{Corridor construction algorithm :$CorridorInfo=Cconstruction(\mathcal{M},P_{i})$}  
    \LinesNumbered
    \KwIn{$\mathcal{M}$: map information\newline
        $P_{i}$: current path point
    }
    \KwOut{current corridor information:$P_{ci,right}^{{H_k}}$, $P_{ci,left}^{{H_k}}$, $\theta_{pi}^{k}$, $P_i$}  
    \textbf{Initialize:} $\Delta_L$,$\Delta_{L0}$, $L_{max}$, $N_c$, $L=[0,0,0,0]$, \\
    \quad\quad\quad\quad\ $AddVec=[1,-1,-1,1]$,\\
    \quad\quad\quad\quad\ $ExpandFlag=[true,true,true,true]$,\\
    \quad\quad\quad\quad\ $CorridorAreaMax=0$,\\
    \quad\quad\quad\quad\ $P_{ci,left}^{{H}}=[-\Delta_{L0},-\Delta_{L0}]$,\\
    \quad\quad\quad\quad\ $P_{ci,right}^{{H}}=[\Delta_{L0},\Delta_{L0}]$\\
    \For{$k \in \{1,2,...,N_c\}$}{
    $\theta_{pi}^{k}=\pi/2*(k-1)/N_c$;\\
    $P_{temp}=[y_{ci,right}^{{H}},x_{ci,left}^{{H}},y_{ci,left}^{H},x_{ci,right}^{H}]$;\\
    \While{\textbf{true}}{
        \If{All elements in ExpandFlag are false }{
            \textbf{break;}
        }
        \For{$j \in \{1,2,3,4\}$}{
            \If{$ExpandFlag[j]==false$}{
            \textbf{continue}
            }
            $L[j]=L[j]+\Delta_L$\\
            $P_{temp}[j]=P_{temp}[j]+\Delta_L*AddVec[j]$;\\
            $V^{H_k}_{ccs}=updateVertex(P_{temp})$;\\
            \If{\\!$IsValid(P_i,V^{H_k}_{Css}[j],V^{H_k}_{Css}[j+1],\theta_{pi}^{k}, \mathcal{M})$  \rm\textbf{and} $L[j]>L_{max}$}{
            $SpandFlag[j]=false$;
            }
        }
    }
    $CA=calculateCorridorArea(P_{temp})$;\\
    \If{$CA>CorridorAreaMax$}{
        $CorridorAreaMax=CorridorArea$;\\
        $P_{ci,right}^{{H_k}}=Vertex[1]$;\\
        $P_{ci,left}^{{H_k}}=Vertex[3]$;\\
        $CorridorInfo=[P_i,P_{ci,left}^{{H_k}},P_{ci,right}^{{H_k}},\theta_{pi}^{k}]$;
    }
    }
    \algorithmicreturn{\ $CorridorInfo$} 
\end{algorithm}

So far, the corridor has been constructed. Next, we will introduce how to use the obtained corridor information to constrain the robot in corridor. When the robot is located in corridor $i$, the robot’s position is transformed into the local coordinate system of the current corridor:

\begin{equation*}
\boldsymbol{\eta}^{H} = \boldsymbol{R}^T(\theta_{pi}^{k})(\boldsymbol{\bar\eta}-\boldsymbol{P}_i)   \tag{8}
\end{equation*}

Then, the robot’s corridor constraint is constructed as the following inequality:

\begin{equation*}
\textbf{A}\boldsymbol{\eta}^{H}-\textbf{b}_i<\textbf{0}   \tag{9}
\end{equation*}
where $\textbf{A}$ and $\textbf{b}_i$ are given as:
\begin{equation*}
\textbf{A}= \begin{bmatrix}1& 0\\ -1& 0\\ 0& 1\\ 0& -1\end{bmatrix}, \textbf{b}_i= \begin{bmatrix}x_{ci,right}^{H}-\Delta_s\\ -x_{ci,left}^{H}-\Delta_s\\ y_{ci,right}^{H}-\Delta_s\\ -y_{ci,left}^{H}-\Delta_s\end{bmatrix} \tag{10}
\end{equation*}
where $\Delta_s>0$ is the inward contraction distance.


\begin{algorithm}
    \caption{Sequential corridors construction algorithm:\ $SCconstruction(P,\mathcal{M})$ }  
    \LinesNumbered
    \KwIn{$P$: path points \newline
        $\mathcal{M}$: map information
    }
    \KwOut{$Corridors$: corridors information}  
    \textbf{Initialize:}\ $N_p=length(P)$;\\
    \quad\quad\quad\quad\ $Corridors = [\ ]$;\\
    $Corridors.append(Cconstruction(P[1],\mathcal{M}))$;\\
    \For{$i \in \{2,3,...,N_p\}$}{
        \If{$isNotInLastCorridor(P[i],Corridors[-1])$}{
            $i=i-1$;\\
            $Corridors.append(Cconstruction(P[i],\mathcal{M}))$;\\
        }
        \If{i==N}{
            $Corridors.append(Cconstruction(P[i],\mathcal{M}))$;
        }
    }
    \algorithmicreturn{\ $Corridors$} 
\end{algorithm}

\begin{algorithm}
    \caption{Improved sequential MPC based navigation algorithm:$ISMPC(Corridors,\mathcal{O},\eta,\eta_d,\mathcal{M})$}  
    \LinesNumbered
    \KwIn{$Corridors$: sequential corridors information \newline
        $\mathcal{O}$: dynamic obstacles information \newline
        $\boldsymbol{\eta}$: robot pose\newline
        $\boldsymbol{\eta}_d$: goal pose\newline
        $\mathcal{M}$: map information
    }
    \KwOut{$\nu_d$: desired velocity}  
    \textbf{Initialize:} $M=length(Corridors)$,$d_{max}$,$DuringTime$\\
    \While{\textbf{not} $reachGoal(\boldsymbol{\eta},\boldsymbol{\eta}_d,d_{max})$}{
        $\boldsymbol{\eta}_{cg1},\boldsymbol{\eta}_{cg2}, currentCorridor = getCurrentCorridor(\boldsymbol{\eta},Corridors)$;\\
        $OT = predictObstacleTrajectory()$;\\
        $\nu_d = solveMPC(\eta_{cg},currentCorridor,OT)$;\\
        \If{$isTimeTriggered(DuringTime)$}{
            $Corridors=SCconstruction(P,\mathcal{M})$;
        }
    }
    \algorithmicreturn{\ $\nu_d$} 
\end{algorithm}

\subsection{Dynamic Control Barrier Function Constraints}
Although corridor construction can be initiated at regular intervals, the extensive time required for path searching diminishes the effectiveness of corridor constraints in evading dynamic obstacles. This necessitates the real-time formulation of constraints to effectively manage dynamic obstacles. To this end, the Dynamic Control Barrier Function (D-CBF) [16] is integrated into the ISMPC framework. For simplicity, dynamic obstacles are modeled as circular occupation regions. The predicted position of the $i$-th dynamic obstacle at the $k$-th step is denoted as $\boldsymbol{\eta}_{oi,k}$, and the system’s safety set can be expressed by the super-level set of a continuously differentiable function $h_{i,k}(\boldsymbol{\eta}, \boldsymbol{\eta}_{oi,k})$:
\begin{equation*}\mathcal{C}=\{\boldsymbol{\eta}\in \mathcal{X}:h_{i,k}(\boldsymbol{\eta}, \boldsymbol{\eta}_{oi,k})\geq 0\} \tag{11}\end{equation*}
Function $h_{i,k}$ serves as a dynamic control barrier function, if it is satisfied if for all $\boldsymbol{\eta} \in \mathcal{D}$, there exist extended K-class functions such that the following condition holds:
\begin{equation*}\exists \boldsymbol{\nu}\ s.t.\dot{h}_{i,k}\geq-\gamma(h_{i,k}), \gamma\in \mathcal{K}_{\infty} \tag{12}\end{equation*}
The safe set $\mathcal{C}$ is forward invariant and asymptotically stable if the condition (12) is satisfied [16]. Therefore, the control input $\boldsymbol{\nu}$ should satisfy the constraint (12) to assure the safety of navigation system. 

The derivative of $h$ can be calculated as:
\begin{equation*}\dot{h}_{i,k}=\frac{\partial h_{i,k}}{\partial \boldsymbol{\eta}}\boldsymbol{g}(\boldsymbol{\eta},\boldsymbol{\nu})+\frac{\partial h_{i,k}}{\partial \boldsymbol{\eta}_{oi,k}}\dot{\boldsymbol{\eta}}_{oi,k}\tag{13}\end{equation*}
where $\dot{\boldsymbol{\eta}}_{oi,k}$ is the velocity vector of dynamic obstacles.

In the context of discrete-time systems, (12) can be reformulated as:
\begin{equation*}\Delta h_{i,k+1} \geq-\Delta T\gamma h_{i,k}, \gamma > 0\tag{14}\end{equation*}
where $\Delta h_{i,k+1}=h_{i,k+1}-h_{i,k}$, $\Delta T$ is the sample time, which is equal to sample time of ISMPC . 


\subsection{Sequential MPC based Navigation Framework}
The dual-layered safety architecture comprising static corridor constraints and D-CBFs have been constructed above. Next, the sequential MPC scheme is given in Algorithm 3: (1) For the input corridor sequence, determine the current corridor information and its index where the robot is located, and construct the corridor constraints. (2) Predict the obstacle trajectory based on the current speed of the obstacle, and construct the D-CBF constraints. (3) Obtain the next target point based on the corridor index, and construct the optimization objective function.
 

         

Once the current corridor, the current target point $\boldsymbol{\eta}_{cg1}$ and next target point $\boldsymbol{\eta}_{cg2}$, and the trajectory of dynamic obstacles are obtained, combined with the MPC framework, the ISMPC scheme can be described as follows:
\begin{align}
\min\limits_{\boldsymbol{\nu}_{k:k+N-1}} & \quad J(t)=\boldsymbol{J}_N(\boldsymbol{\eta}_{k+N\vert k})+ \sum\limits_{k=0}^{N-1}\boldsymbol{J}_k(\boldsymbol{\eta}_{k}, \boldsymbol{\nu}_{k})\tag{15a} \\
\quad\quad\mathrm{s}.\mathrm{t}. 
&\quad\boldsymbol{\eta}_{k+1}=\boldsymbol{g}(\boldsymbol{\eta}_k,\boldsymbol{\nu}_k)\tag{15b}\\ 
&\quad\boldsymbol{\nu}_{k}\in \mathcal{U} \tag{15c}\\
&\quad\textbf{A}\boldsymbol{\eta}^{H}_{k+1}-\textbf{b}_i<\textbf{0} \tag{15d-1}\\
&\quad\Delta h_{j,k+1} \geq-\Delta T\gamma h_{j,k} \tag{15d-2}\\
&\quad\boldsymbol{\eta}_{0}= \boldsymbol{\eta}_{t}\tag{15e}
\end{align}
where  $\boldsymbol{J}_N(\boldsymbol{\eta}_{k+N\vert k})=||\boldsymbol{\eta}-\boldsymbol{\eta}_{cg1}||_{Q1}+||\boldsymbol{\eta}-\boldsymbol{\eta}_{cg2}||_{Q2}$ and $\boldsymbol{J}_k(\boldsymbol{\eta}_{k}, \boldsymbol{\nu}_{k})=||\boldsymbol{\eta}-\boldsymbol{\eta}_{cg1}||_{Q1}+||\boldsymbol{\eta}-\boldsymbol{\eta}_{cg2}||_{Q2}+||\boldsymbol{\nu}_k||_R+||\boldsymbol{\nu}_{k+1}-\boldsymbol{\nu}_{k}||_S$, $k=0, \ldots, N-1$, $i=0,\ldots, N_c-1$, $N_c$ is the number of corridors, $j=1,2,...,N_o$, $j$ and $N_o$ are the index and number of dynamic obstacles respectively.

\section{Simulation result}
The numerical simulations were performed on a computational platform equipped with an Intel i7-14650HX CPU (16 GB RAM), using the Gazebo 11 simulator integrated with ROS1 Noetic. The fundamental simulation parameters are systematically summarized in Table I. The OCP formulated in Equations (15a)-(15e) was solved through the IPOPT solver implemented within the CasADi(C++) framework. To rigorously validate the algorithm's enhanced performance in unstructured and dynamic environments, three distinct experimental cases were designed. Simulation results are shown in Figure 5-Figure 10 and Table II. 

\begin{table}[htbp]
\caption{PARAMETERS VALUE}
\begin{center}
\begin{tabular}{c|c|c}
\Xhline{1.5pt}
\textbf{Parameter} & \textbf{Description} & \textbf{Value} \\
\Xhline{1pt}
$u_{max}$& Maximum linear velocity & 1.0 \\
\hline
$u_{min}$& Minimum linear velocity & 0.05 \\
\hline
$r_{max}$& Maximum angular velocity & 1.5 \\
\hline
$\Delta_{L}$& Length of single corridor extension & 0.1 \\
\hline
$\Delta_{s}$& Inward contraction distance & 0.3 \\
\hline
$L_{max}$& Maximum Length of corridor extension  & 8.0 \\
\hline
$N_{c}$& The number of expansion directions& 10 \\
\hline
$Q_1$& Weight matrix of the first goal & diag{[20,20,0.1]} \\
\hline
$R_1$& Weight matrix of the second goal & diag{[5,5,0.05]} \\
\hline
$R$& Weight matrix of control inputs & diag{[0.1,0.1]} \\
\hline
$N_{s}$& Predict step  & 10 \\
\hline
$t_{s}$& Sample time  & 0.1 \\
\hline
$R_o$& Radius of dynamic obstacles & 0.8 \\
\hline
$\gamma$& The coefficients of D-CBF & 5 \\
\Xhline{1.5pt}
\end{tabular}
\label{tab1}
\end{center}
\end{table}

\noindent \textbf{Case 1: Structured Obstacle-Dense Environment.}

The environment with many static and structured obstacles is constructed to demonstrate the workflow and effectiveness of the method proposed in this paper. The results are shown in Figure 5 and Figure 6. As shown in Figure 5(a), the green rectangular boxes generated by MDSRC ensure that the AMR avoids collision with all static obstacles. A comparative analysis of path planning outcomes between SMPC in [17] and ISMPC in this paper is presented in Figures 5(a) and 5(b). The results demonstrate that the proposed methodology generates trajectories with shorter path lengths and reduced redundant maneuvering compared to conventional approaches. This performance enhancement is primarily attributed to two algorithmic modifications: (1) direct utilization of A*-derived waypoints as intermediate targets, eliminating computational overhead associated with intersection centroid calculations, and (2) incorporation of dual subsequent nodal points rather than singular goal within the Model Predictive Control framework. The implemented strategy effectively optimizes path smoothness while maintaining computational efficiency, as evidenced by the comparative trajectory profiles. Figure 6 shows that the speed and angular velocity of the AMR are each kept within the pre-established boundaries of 1 m/s and 1.5 rad/s. In addition, the differential driven mobile robot can perfectly track the desired speed due to the smooth velocity commands.

\begin{figure}
    \centering
    \includegraphics[width=1\linewidth]{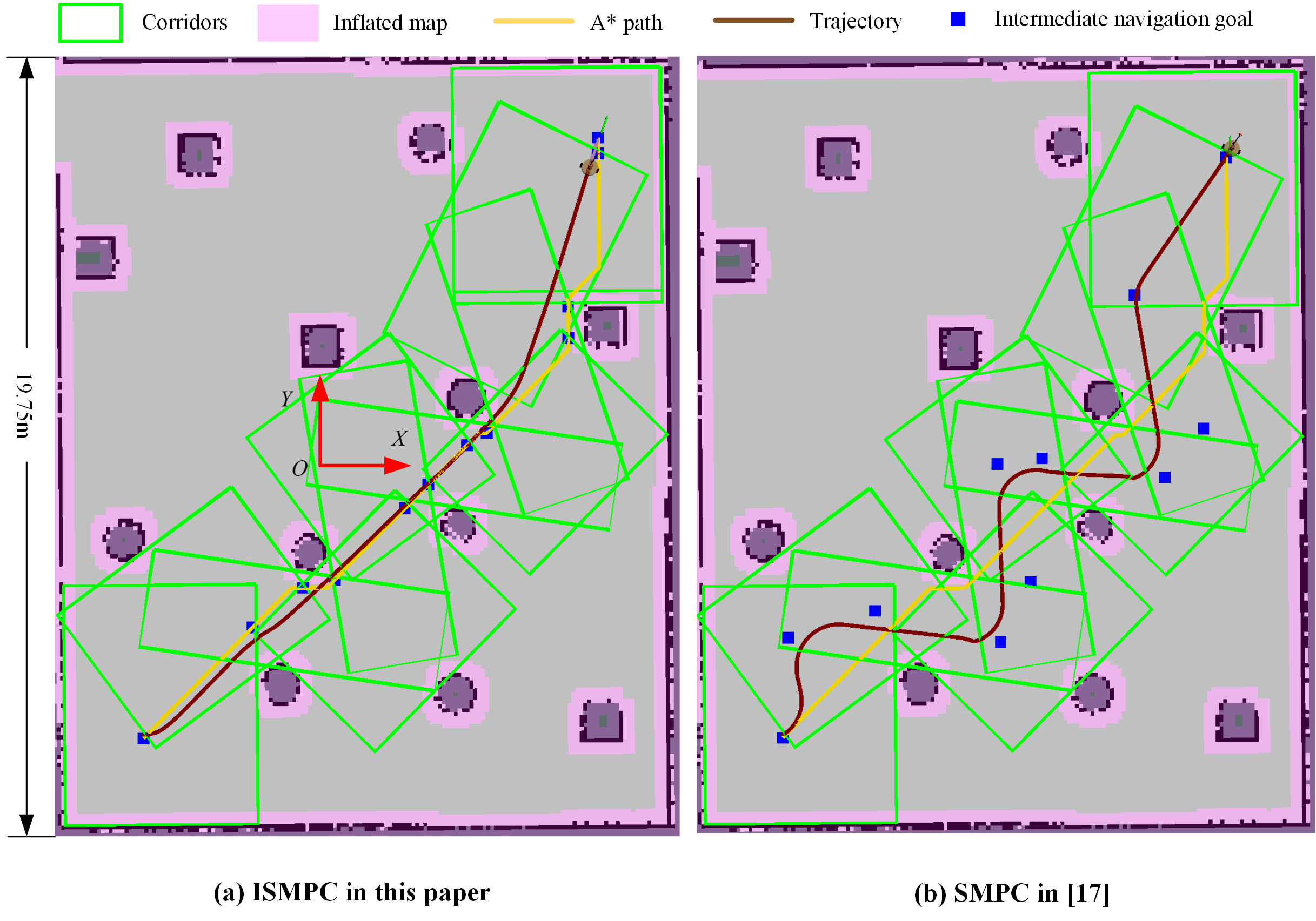}
    \caption{Generated corridors and real trajectory of the AMR in Case 1. The green points are the seed points for corridor generation and also serve as the intermediate target points for navigation.}
    \label{fig:enter-label}
\end{figure}

\begin{figure}
    \centering
    \includegraphics[width=1\linewidth]{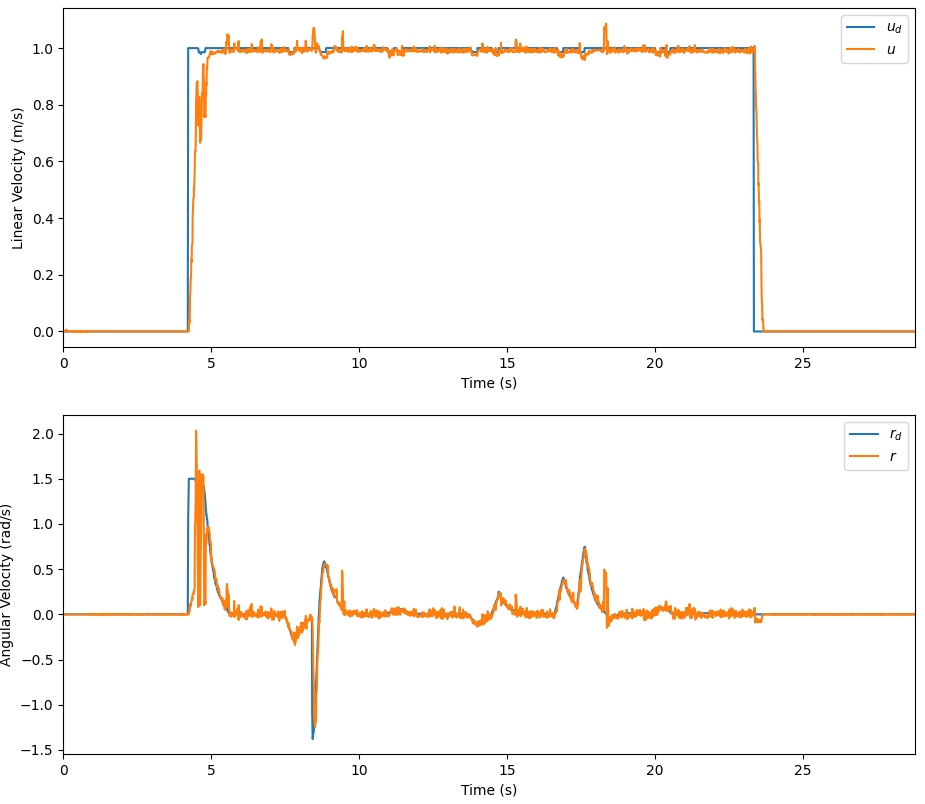}
    \caption{Velocities of the mobile robot.}
    \label{fig:enter-label}
\end{figure}

\begin{figure}
    \centering
    \includegraphics[width=1\linewidth]{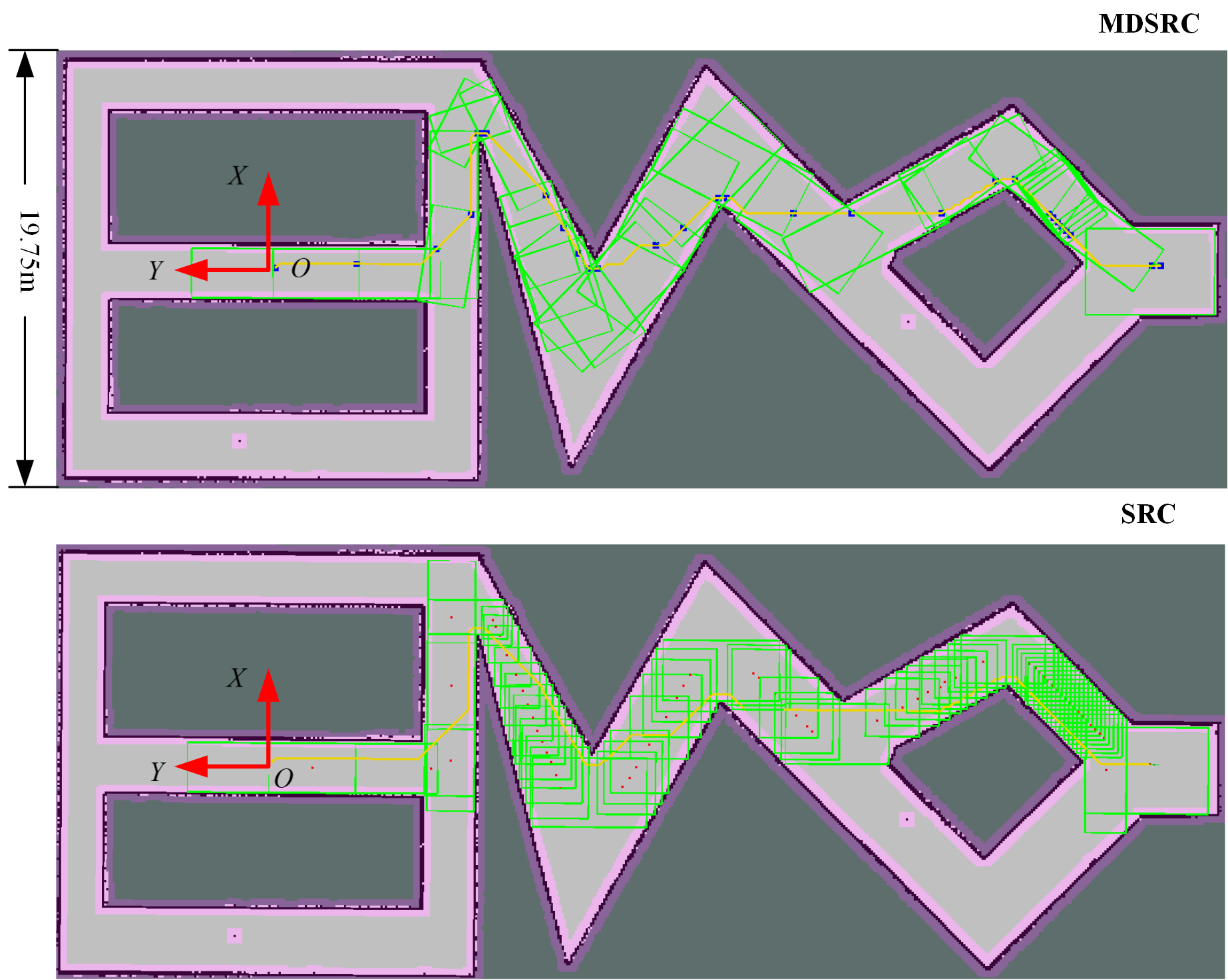}
    \caption{Generated corridors in Environment 1 of Case 2.}
    \label{fig:enter-label}
\end{figure}

\begin{figure}
    \centering
    \includegraphics[width=1\linewidth]{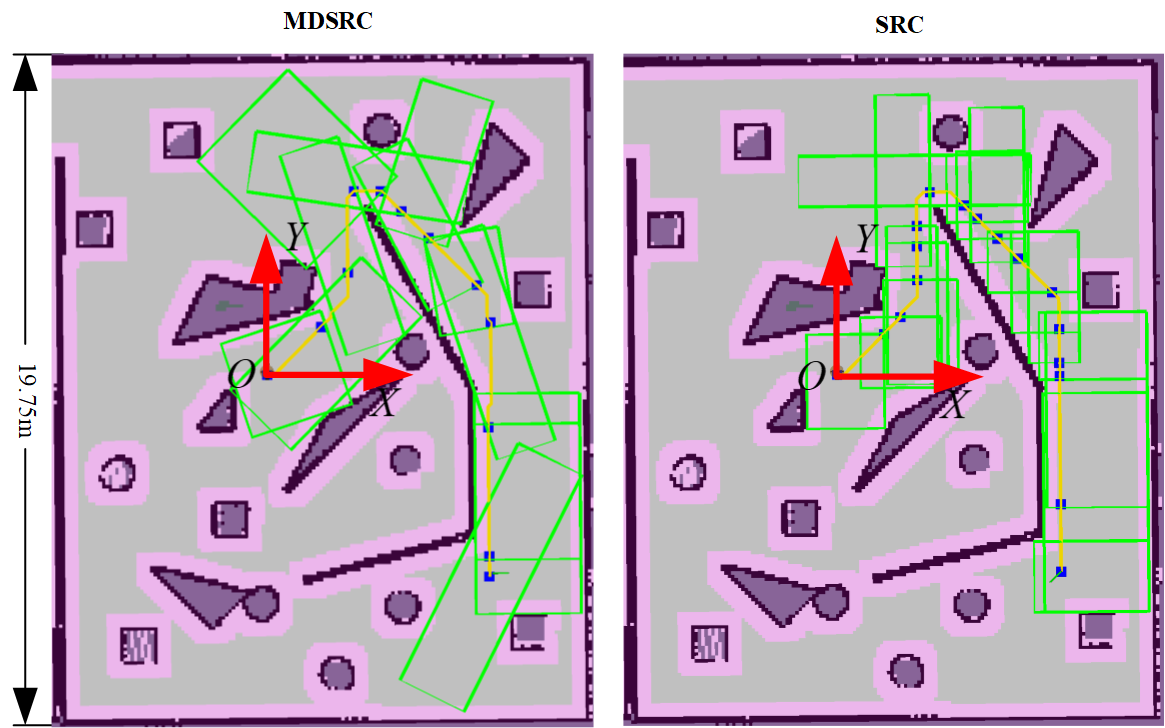}
    \caption{Generated corridors in Environment 2 of Case 2.}
    \label{fig:enter-label}
\end{figure}

\noindent \textbf{Case 2: Unstructured Obstacle-Dense Environment.}

To validate the efficacy of the MDSRC generation method in unstructured environments characterized by irregular obstacle configurations, a comparative analysis was conducted against conventional Static Rectangular Corridor (SRC) [13] and Fast Static Rectangular Corridor (FSRC) [14] methods. As illustrated in Figures 7-8 and Table II, MDSRC demonstrates superior adaptability in geometrically complex scenarios where obstacles lack axial alignment. Unlike SRC and FSRC, which generate axis-aligned rectangular corridors necessitating redundant spatial segmentation in diagonal passages (Figures 7-8), the proposed method produces fewer yet larger safety regions that better conform to irregular environmental constraints. The quantitative evaluation results are displayed in Table II, in which, $NC$, $MA$(m$^2$) and $TS$(s) are the total number, average area 
and  total generation time of corridors respectively.

\begin{table}[htbp]
\caption{COMPARISON RESULTS}
\begin{center}
\begin{tabular}{c|c|c|c|c|c|c}
\Xhline{1.5pt}
\multirow{2}{*}{\textbf{Method}}&\multicolumn{3}{c|}{\textbf{Environment 1}}&\multicolumn{3}{c}{\textbf{Environment 2}} \\
\cline{2-7} 
& \textbf{\textit{NC}}& \textbf{\textit{MA}}& \textbf{\textit{TS}} & \textbf{\textit{NC}}& \textbf{\textit{MA}}& \textbf{\textit{TS}} \\
\Xhline{1pt}
SRC& 52 & 6.94 & 0.002 & 17 & 7.59 & 0.001 \\
\hline
FSRC& 52 & 6.94 & 0.001 & 17 & 7.59 & 0.0002 \\
\hline
MDSRC& 30 & 12.80 & 0.004 & 12 & 11.92 & 0.002 \\
\Xhline{1.5pt}
\end{tabular}
\label{tab1}
\end{center}
\end{table}

\begin{figure}
    \centering
    \includegraphics[width=1\linewidth]{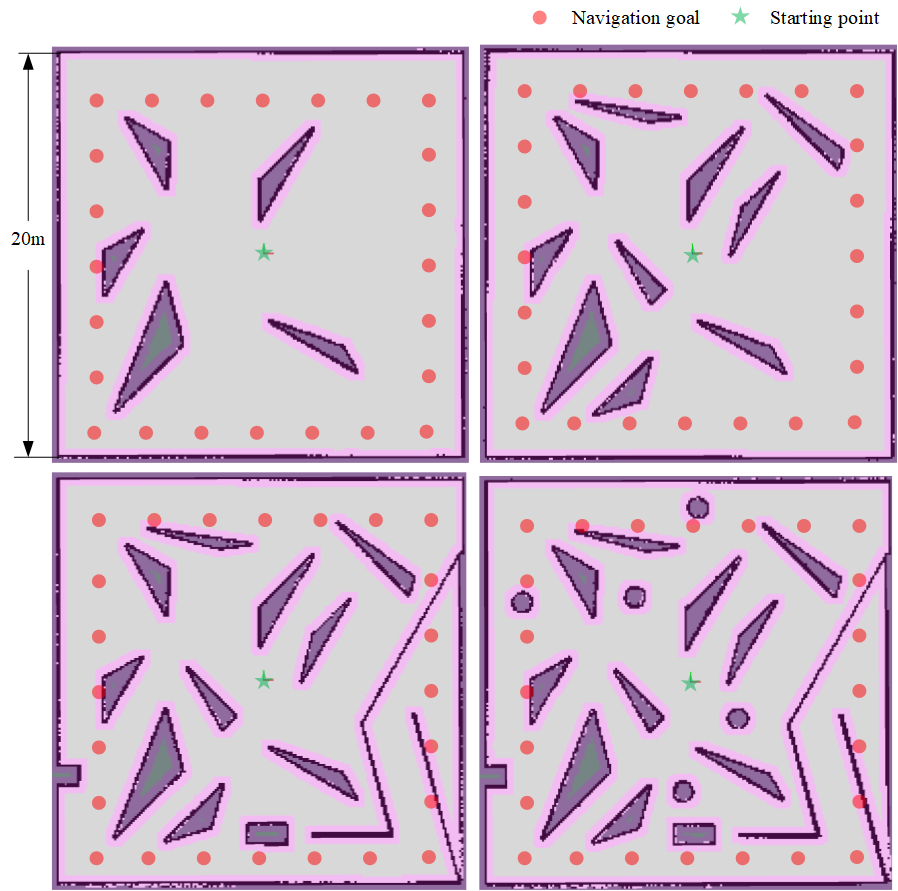}
    \caption{Scenarios with different numbers of obstacles in Environment 3 of Case 2. The obstacle numbers are set as 5, 10, 15, and 20 respectively, and 24 target points are chosen in each scenario.}
    \label{fig:enter-label}
\end{figure}

\begin{figure}
    \centering
    \includegraphics[width=1\linewidth]{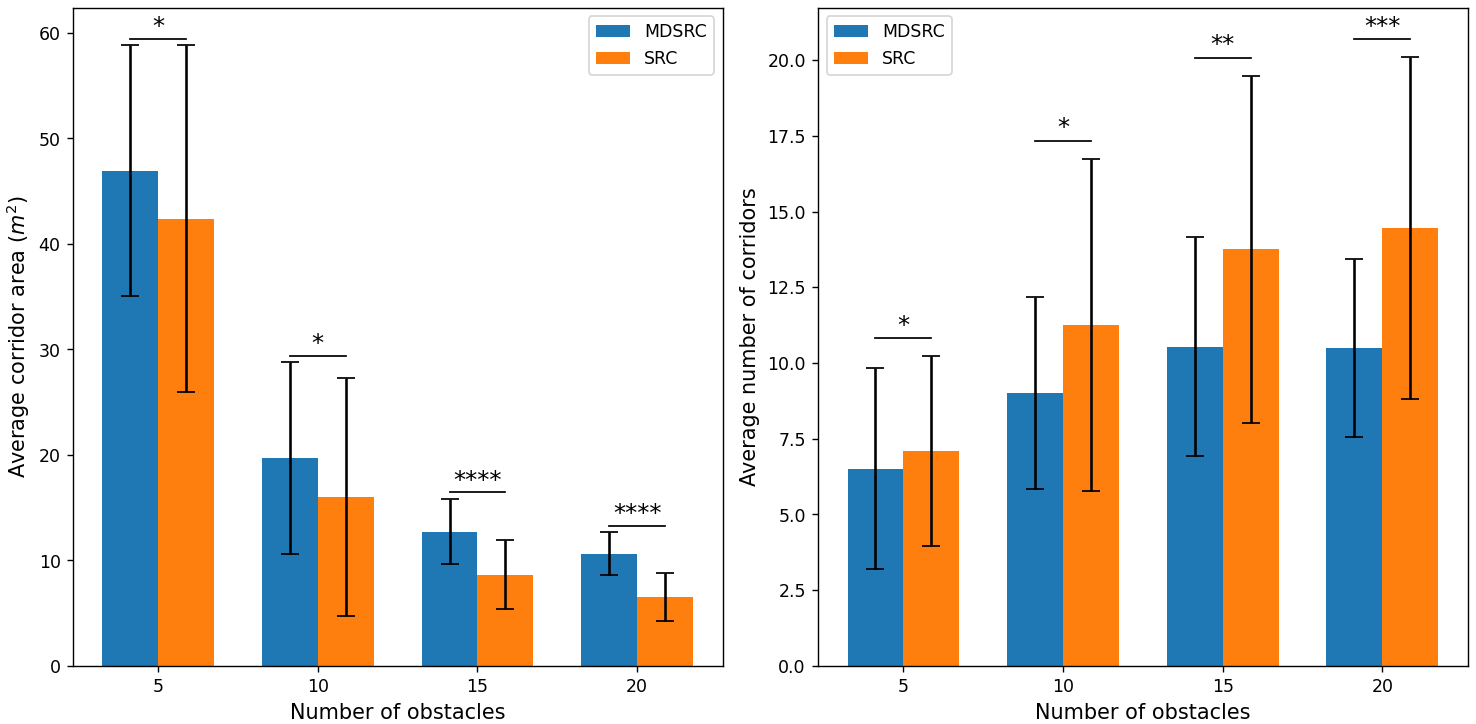}
    \caption{Variations of average corridor area and total number of corridors with obstacle quantity. The average corridor area and number for $N$ obstacles are the means of areas and quantities across all target - point - generated corridors. Error bars represent data variance. Asterisks denote statistical significance levels as follows: *$p>0.05$ (ns, not significant); **$0.01<p\le 0.05$; ***$0.001 < p\le 0.01$; ****$p\le 0.001$.}
\end{figure}
\setlength{\textfloatsep}{5pt plus 2pt minus 2pt}
Quantitative evaluations confirm that MDSRC achieves enhanced spatial coverage per corridor while reducing fragmentation compared to benchmark methods. As shown in Table II, the average number of rectangular corridors decreased by 35.86$\%$, while the average corridor area increased by 41.05$\%$. Although marginally slower in computation than FSRC, the algorithm maintains computational efficiency within acceptable thresholds for real-time navigation tasks. These results collectively validate the robustness of MDSRC in enabling reliable path planning for dense, unstructured environments with non-uniform obstacle distributions.

To further assess the adaptability of the proposed algorithm under varying obstacle numbers, 24 points are randomly selected on a 20m×20m map for corridor generation experiments using MDSRC and SRC. The map and selected points are shown in Figure 9, with results presented in Figure 10. The figure reveals that MDSRC generates corridors with a larger average area and fewer in number compared to SRC, and this advantage amplifies as obstacles increase. Larger areas facilitate agent obstacle avoidance, while fewer corridors reduce switching times and control signal fluctuations. Thus, MDSRC outperforms SRC in overall performance.

\noindent \textbf{Case 3: Dynamic Multi-Agent Navigation.}
\begin{figure}
    \centering
    \includegraphics[width=1\linewidth]{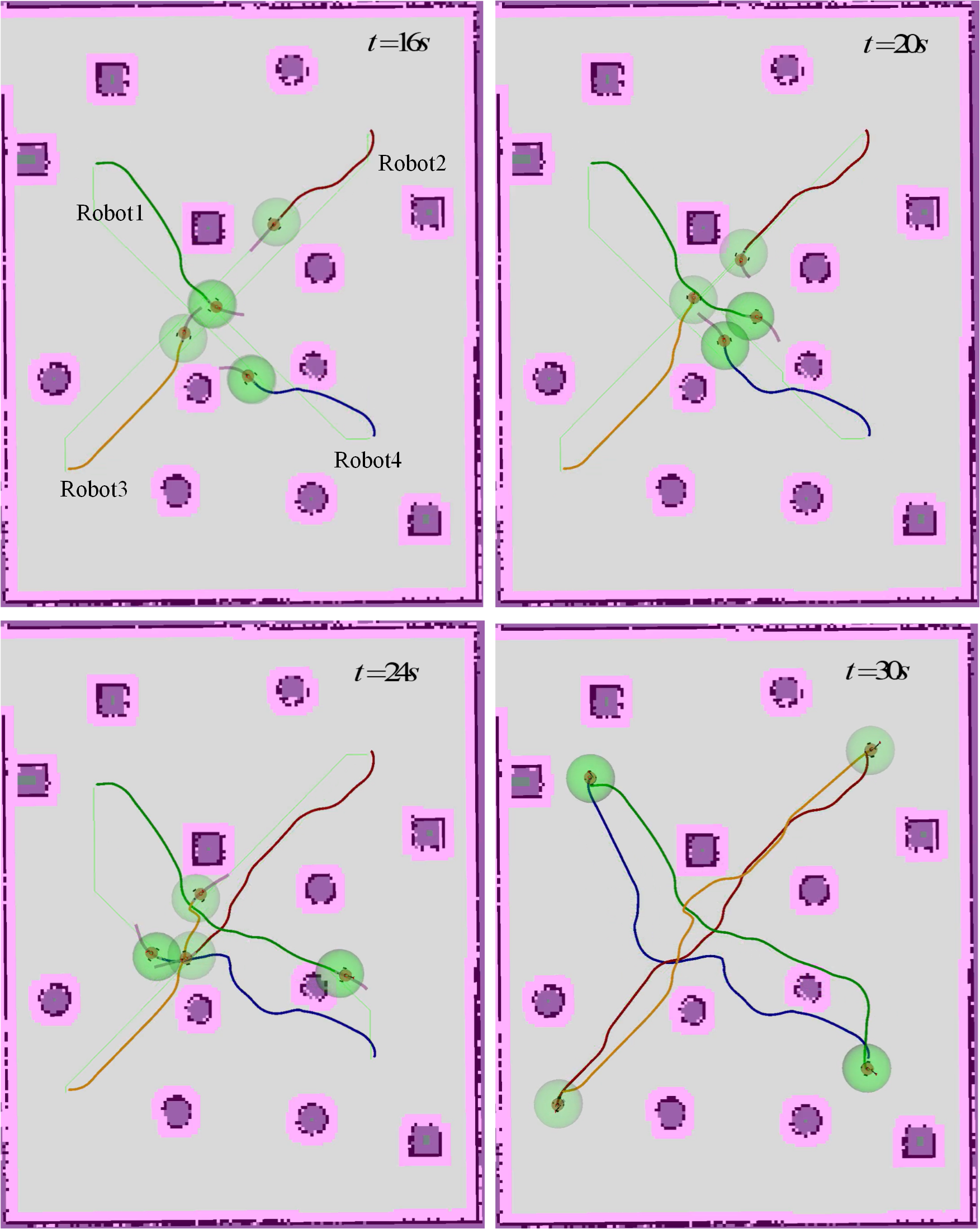}
    \caption{Temporal snapshots of dynamic obstacle avoidance behaviors using ISMPC in Case 3. The green ellipsoidal area denotes the collision range. Purple lines are predicted trajectories, and thick lines of other colors represent robot's trajectories. Corridors are omitted here for better visualization.}
    \label{fig:enter-label}
\end{figure}
The proposed methodology was further validated in a dynamic environment involving four mobile robots with interchanged target positions to rigorously assess its collision avoidance capabilities under time-varying conditions. For efficient algorithm execution, the speeds and positions of all robots are shared in real time to enable trajectory prediction. As demonstrated in Figure 11, all agents successfully navigated to designated targets while maintaining collision-free trajectories despite emergent path conflicts. These results collectively confirm the framework's capacity to negotiate complex spatiotemporal conflicts through anticipatory trajectory modulation while ensuring operational safety in dense dynamic environments.

\section{Conclusion}
This study proposes an improved sequential model predictive control framework. It combines dynamic control barrier functions with safety corridor constraints to systematically handle the challenges of avoiding both static and dynamic obstacles for autonomous mobile robots, ensuring collision-free navigation in complex environments. A major innovation is the Multi-directional Safety Rectangle Corridor (MDSRC) generation method. It expands the feasible solution regions adaptively while keeping computational efficiency. Numerous Gazebo-based simulations prove that the framework can tackle complex navigation tasks in dense obstacle setups. In the future, the feasibility and convergence conditions of MPC will be explored, and the pedestrian trajectory prediction module will be incorporated to realize social navigation in complex pedestrian environments.


\end{document}